\documentclass[a4paper]{article}
\usepackage[T1]{fontenc}
\usepackage{graphicx,verbatim}
\usepackage{svg}
\usepackage[colorlinks]{hyperref}
\usepackage{color}

\usepackage{tabularx} 
\usepackage{multirow} 
\usepackage[para]{threeparttable}  
\usepackage{makecell}
\usepackage{subcaption} 
\usepackage{overpic}  
\usepackage{xcolor}
\usepackage{rotating} 
\usepackage{siunitx} 
\usepackage{orcidlink}
\usepackage[misc,geometry]{ifsym}

\usepackage{array}
\newcommand{\PreserveBackslash}[1]{\let\temp=\\#1\let\\=\temp}
\newcolumntype{C}[1]{>{\PreserveBackslash\centering}p{#1}}
\newcolumntype{R}[1]{>{\PreserveBackslash\raggedleft}p{#1}}
\newcolumntype{L}[1]{>{\PreserveBackslash\raggedright}p{#1}}

\usepackage{amsmath}		
\usepackage{amssymb}
\usepackage{amsfonts}		
\DeclareMathOperator{\E}{\mathbb{E}}
\DeclareMathOperator{\Proba}{Pr}
\DeclareMathOperator*{\argmin}{argmin}
\DeclareMathOperator*{\argmax}{argmax}

\begin{document}
\title{Anomaly Detection by Clustering DINO Embeddings using a Dirichlet Process Mixture}
%
\date{}
\author{
Nico Schulthess\textsuperscript{(\Letter)}\orcidlink{0000-0002-3421-5638} and Ender Konukoglu\orcidlink{0000-0002-2542-3611} \vspace{0.3cm} \\
Computer Vision Lab, ETH Zurich, Zurich, Switzerland\\
\href{mailto:nico.schulthess@vision.ee.ethz.ch}{\texttt{nico.schulthess@vision.ee.ethz.ch}}
}

\maketitle              
\begin{abstract}
In this work, we leverage informative embeddings from foundational models for unsupervised anomaly detection in medical imaging.
For small datasets, a memory-bank of normative features can directly be used for anomaly detection which has been demonstrated recently.
However, this is unsuitable for large medical datasets as the computational burden increases substantially.
Therefore, we propose to model the distribution of normative DINOv2 embeddings with a Dirichlet Process Mixture model (DPMM), a non-parametric mixture model that automatically adjusts the number of mixture components to the data at hand.
Rather than using a memory bank, we use the similarity between the component centers and the embeddings as anomaly score function to create a coarse anomaly segmentation mask.
Our experiments show that through DPMM embeddings of DINOv2, despite being trained on natural images, achieve very competitive anomaly detection performance on medical imaging benchmarks and can do this while at least halving the computation time at inference.
Our analysis further indicates that normalized DINOv2 embeddings are generally more aligned with anatomical structures than unnormalized features, even in the presence of anomalies, making them great representations for anomaly detection.
The code is available at \url{https://github.com/NicoSchulthess/anomalydino-dpmm}.\footnote{Paper accepted at MICCAI 2025}

\subsubsection*{Keywords}
Unsupervised Anomaly Detection $\cdot$ Dirichlet Process Mixture Model $\cdot$  Foundation Model

\end{abstract}
\section{Introduction}
Anomaly detection focuses on identifying samples that deviate from the norm, such as the detection of lesions in medical imaging or defects in industrial inspection.
Here, the norm is defined by normal example samples, for instance by images of healthy volunteers in medical contexts or intact objects in industrial settings.
Of particular interest is the unsupervised setting, where no anomalous samples are used during training.
This avoids introducing unwanted biases towards certain types of anomalies, does not introduce any potential class imbalances, and alleviates the need for annotating anomalous samples for each modality and anatomy.

One line of work to tackle unsupervised anomaly detection~(UAD) are recon\-struc\-tion-based approaches, where a generative model is trained to reconstruct normal images.
The underlying hypothesis is that since the model is trained to reconstruct normal images well, it will do a poor job in anomalous regions and the reconstruction error will highlight such regions. 
Popular choices for the reconstruction model are adaptations of autoencoders~\cite{chen2018unsupervised,Lu2024Heterogeneous}, variational autoencoders~\cite{Zimmerer2019VAE,chen2020unsupervised,chen2021normative}, masked autoencoders~\cite{Rashmi2024Ano-SwinMAE}, generative adversarial networks~\cite{Schlegl2019fAnoGAN}, and denoising diffusion probabilistic models~\cite{Wyatt2022AnoDDPM}.
Instead of reconstructing images, a feature reconstruction objective can be used in a knowledge distillation setup where a student network is trained on normal data to predict similar features to a pretrained teacher network~\cite{Liu2024SkipSTJournal}.
Then, the embeddings are assumed to differ between the student and teacher networks for anomalous data while they are similar for normal data.

Another line of work aims to model the distribution of the features extracted from normal samples.
One choice is to use normalizing flow models to map the normal patch feature distribution to a tractable distribution~\cite{Gudovskiy2022CFLOW-AD}.
However, normalizing flow models can assign high probabilities to anomalous samples and should thus be used with caution for anomaly detection~\cite{Kirichenko2020Normalizing,Nalisnick2018Do}.
SPADE~\cite{Cohen2021SPADE} constructs a memory bank of normal embeddings and uses the average distance to k nearest neighbors as an anomaly score for a given sample.
PaDiM~\cite{Defard2021PaDiM} creates a probabilistic representation of the features by modeling feature distribution for each pixel by a multivariate Gaussian.
PatchCore~\cite{Roth2022PatchCore} and ProtoAD~\cite{Huang2024ProtoAD} employ a memory bank as in SPADE, however, both methods compress the memory bank to reduce its size using a greedy coreset selection and a nonparametric hierarchical clustering, respectively.
AnomalyDINO~\cite{Damm2024AnomalyDINO} stores patch features from DINOv2~\cite{Oquab2024DINOv2} using only a few normal images to construct a memory bank.

The comparison with large memory banks usually leads to accurate anomaly detections, however at the cost of substantially long runtimes and high memory utilization.
Having small yet expressive memory banks is crucial for anomaly detection on a large scale as it would be required for medical imaging.

\begin{figure}[t]
\centering
\includegraphics[width=\textwidth]{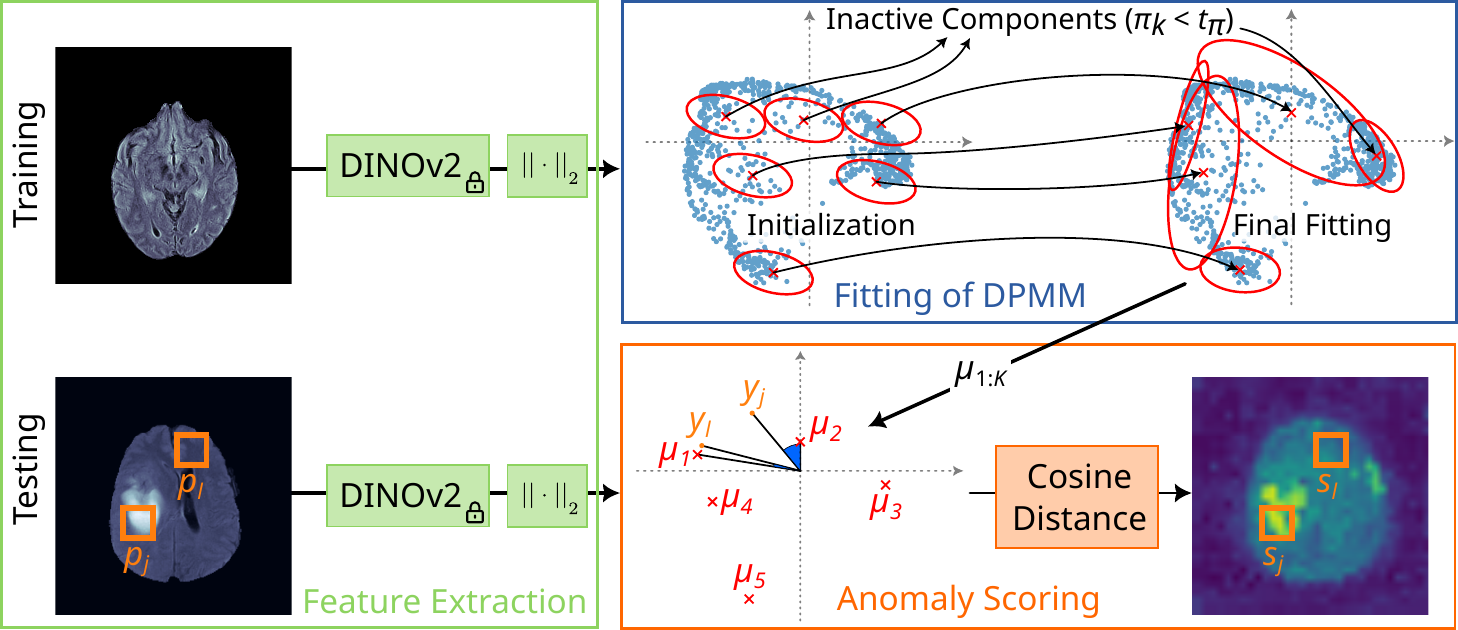}
\caption{
In the training phase, we fit a Dirichlet process mixture model~(DPMM) with Gaussian components~(indicated in red) to the DINOv2 patch embeddings of the normal samples.
After fitting the DPMM to the normal embeddings, we use the cosine distance between each patch feature and the closest component mean to obtain a patch-level anomaly score, which we interpolate for a pixel-wise anomaly map.
}
\label{fig:meth:overview}
\end{figure}

In this work, we aim to effectively compress the memory bank of AnomalyDINO.
We propose to model the patch features using a Dirichlet process mixture model (DPMM)~\cite{Antoniak1974DPMM,Ferguson1983DPMM} with Gaussian component distributions, which is a Gaussian mixture model without a fixed number of components that instead determines the necessary number of components based on the data, allowing for a more flexible, data-driven approach.
We fit the DPMM using a batched expectation maximization algorithm~\cite{Kimura2013Expectation}.
We use the component means as prototypes for the normal distribution and define the anomaly score as the similarity to these prototypes.
For an overview of our approach refer to Fig.~\ref{fig:meth:overview}.

Our framework achieves competitive performance on the BMAD benchmark~\cite{Bao2024BMAD} while being highly efficient to compute thanks to the small number of prototypes in use to model the normative distribution.
Further, our analyses indicate that features from DINOv2 trained on natural images are useful for anomaly detection in medical imaging, especially after normalization.

\section{Methods}
\subsection{Fitting the Dirichlet Process Mixture Model~(DPMM)}\label{sec:meth:train}
Following \cite{Kimura2013Expectation}, we fit the DPMM with Gaussian components using a batched expectation maximization~(EM) approach.
For this, we consider a DPMM truncated to $K$~components.
For sufficiently large $K$, the truncated DPMM is more flexible than a standard Gaussian mixture model as components can vanish in the DPMM if they are not required.
We define the probability density of a sample~$y$ under the DPMM with parameters~$\Phi$ as
\begin{equation}\label{eq:meth:likelihood}
p\left(y | \Phi\right) = \sum_{k=1}^{K} \pi_{k} \mathcal{N}(y | \theta_{k}),
\end{equation}
where $\theta_{k} = (\mu_{k}, \Sigma_{k})$ and the component weights~$\{\pi_{k}\}_{k=1}^{K}$ are defined as
\begin{equation}
\pi_{k} = \begin{cases}
v_{1}, \quad & \text{if } k = 1\\
v_{k} \prod\limits_{j=1}^{k-1} (1 - v_{j}), \quad & \text{otherwise},
\end{cases}
\end{equation}
with $v_{k} \sim \text{Beta}(1, \alpha)$ for $k \in \{1, \dots, K-1\}$, $v_{K} = 1$, and $\alpha$ being the concentration parameter of the Dirichlet process. 
Given a batch of embeddings $y_{1:N}$, we want to estimate the parameters $\Phi = (v_{1:K-1}, \theta_{1:K})$ and $\alpha$ by maximizing
\begin{equation}
Q_{\Phi}(\Phi, \Phi^{(t-1)}) = \E_{x_{1:N} | y_{1:N}, \theta_{1:K}^{(t-1)}, \alpha^{(t-1)}} \left[ \log p(x_{1:N}, y_{1:N}, v_{1:K} | \theta_{1:K}, \alpha) \right]
\end{equation}
  and
\begin{equation}
Q_{\alpha}(\alpha, \alpha^{(t-1)}) = \E_{x_{1:N} | y_{1:N}, \theta_{1:K}^{(t-1)}, \alpha^{(t-1)}} \left[ \log \int p(x_{1:N}, y_{1:N}, v_{1:K} | \theta_{1:K}, \alpha) \,\mathrm{d} v_{1:K} \right]
\end{equation}
with respect to $\Phi$ and $\alpha$, respectively, where $x_{n} \in \{1, \dots, K\}$ is the random variable for the component assignment of the embedding $y_{n}$.

For Gaussian components, we can obtain the update formulas
\begin{align}
\theta_{k}^{(t)} = \left(\mu_{k}^{(t)}, \Sigma_{k}^{(t)}\right) &= \left(\frac{\bar{m}_{k}^{(t)}}{\sum_{j=1}^{K}\bar{p}_{j}^{(t)}}, \quad\frac{\bar{c}_{k}^{(t)}}{\sum_{j=1}^{K}\bar{p}_{j}^{(t)}} - \mu_{k}^{(t)}\mu_{k}^{(t)^{\top}}\right)\\
v_{k}^{(t)} &= \frac{\bar{p}_{k}^{(t)}}{\bar{p}_{k}^{(t)} + \alpha^{(t-1)} - 1 + \sum_{j=k+1}^{K}\bar{p}_{j}^{(t)}},
\end{align}
where $\bar{p}_{k}$, $\bar{m}_{k}$, and $\bar{c}_{k}$ are moving averages of the sufficient statistics defined by
\begin{alignat}{2}\label{eq:meth:moving_average1}
\bar{p}_{k}^{(t)} &= (1 - \gamma_{t}) \bar{p}_{k}^{(t - 1)} &&+ \gamma_{t} \E_{y}\left[\Proba(x = k | y, \Phi^{(t-1)}) \right]\\\label{eq:meth:moving_average2}
\bar{m}_{k}^{(t)} &= (1 - \gamma_{t}) \bar{m}_{k}^{(t - 1)} &&+ \gamma_{t} \E_{y}\left[y \Proba(x = k | y, \Phi^{(t-1)}) \right]\\\label{eq:meth:moving_average3}
\bar{c}_{k}^{(t)} &= (1 - \gamma_{t}) \bar{c}_{k}^{(t - 1)} &&+ \gamma_{t} \E_{y}\left[y y^{\top}\Proba(x = k | y, \Phi^{(t-1)}) \right].
\end{alignat}
Finally, the update formula for the concentration parameter~$\alpha$ is
\begin{equation}
\alpha^{(t)} = \frac{N - 1}{\sum_{k = 1}^{N - 1}\Psi\left( \alpha^{(t - 1)} + 1 + \bar{C}_{k}^{(t)} + \bar{C}_{>k}^{(t)}\right) - \Psi\left( \alpha^{(t - 1)} + \bar{C}_{>k}^{(t)}\right)},
\end{equation}
with $\bar{C}_{k}^{(t)} = B \cdot \bar{p}_{k}^{(t)}$, $\bar{C}_{>k}^{(t)} = B \sum_{j=k+1}^{K} \bar{p}_{j}^{(t)}$, and $B$ being the batch size.
Further details on the derivation of the formulas above can be found in~\cite{Kimura2013Expectation}.

\subsection{Anomaly Scoring}\label{sec:meth:test}
\begin{figure}[t]
\centering
\begin{minipage}[t][\baselineskip][t]{0.46\textwidth}
    \centering
    Normal Samples
\end{minipage}\hspace{1mm}
\begin{minipage}[t][\baselineskip][t]{0.46\textwidth}
    \centering
    Anomalous Samples
\end{minipage}
\begin{subfigure}[c]{0.116\textwidth}
	\begin{minipage}[t][0.7\baselineskip][t]{\textwidth}
		\centering\footnotesize
		Image
	\end{minipage}
	\includegraphics[angle=270, origin=c, width=\textwidth]{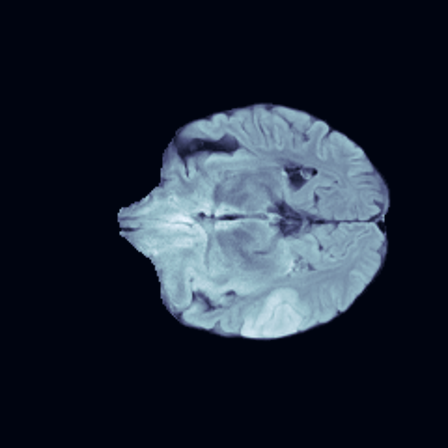}
	\includegraphics[angle=270, origin=c,width=\textwidth]{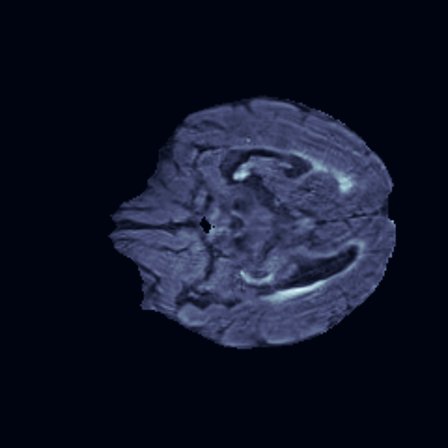}
\end{subfigure}
\begin{subfigure}[c]{0.116\textwidth}
    \begin{minipage}[t][0.7\baselineskip][t]{\textwidth}
		\centering\footnotesize
		Likelihood
	\end{minipage}
	\begin{overpic}[width=\textwidth,percent,angle=270,origin=c]{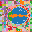}\put(15,30){\includegraphics[scale=1,angle=0, origin=lt]{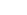}}\end{overpic}
	\begin{overpic}[width=\textwidth,percent,angle=270,origin=c]{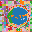}\end{overpic}
\end{subfigure}
\begin{subfigure}[c]{0.116\textwidth}
    \begin{minipage}[t][0.7\baselineskip][t]{\textwidth}
		\centering\footnotesize
		Euclidean
	\end{minipage}
	\begin{overpic}[width=\textwidth,percent,angle=270,origin=c]{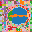}\end{overpic}
	\begin{overpic}[width=\textwidth,percent,angle=270,origin=c]{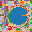}\end{overpic}
\end{subfigure}
\begin{subfigure}[c]{0.116\textwidth}
    \begin{minipage}[t][0.7\baselineskip][t]{\textwidth}
		\centering\footnotesize
		Cosine
	\end{minipage}
	\begin{overpic}[width=\textwidth,percent,angle=270,origin=c]{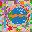}\end{overpic}
	\begin{overpic}[width=\textwidth,percent,angle=270,origin=c]{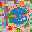}\put(8,4){\includegraphics[scale=1,angle=45]{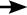}}\end{overpic}
\end{subfigure}\hspace{1mm}
\begin{subfigure}[c]{0.116\textwidth}
    \begin{minipage}[t][0.7\baselineskip][t]{\textwidth}
		\centering\footnotesize
		Image
	\end{minipage}
	\begin{overpic}[width=\textwidth,percent,angle=270,origin=c]{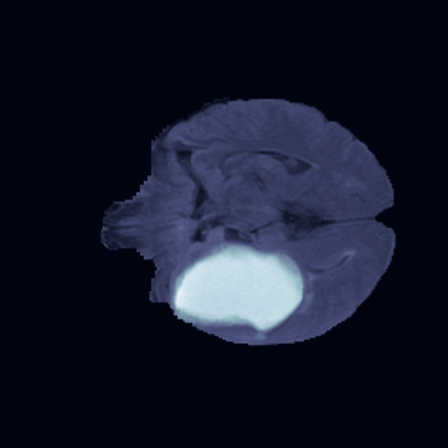}\end{overpic}
	\begin{overpic}[width=\textwidth,percent,angle=270,origin=c]{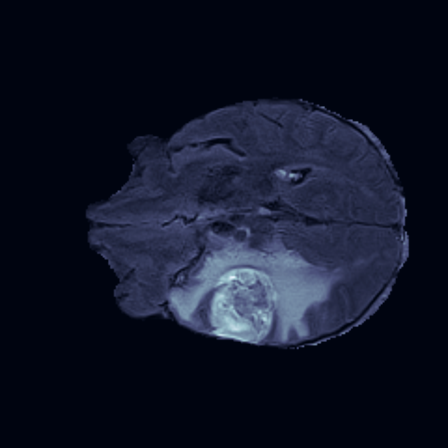}\end{overpic}
\end{subfigure}
\begin{subfigure}[c]{0.116\textwidth}
    \begin{minipage}[t][0.7\baselineskip][t]{\textwidth}
		\centering\footnotesize
		Likelihood
	\end{minipage}
	\begin{overpic}[width=\textwidth,percent,angle=270,origin=c]{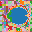}\put(15,30){\includegraphics[scale=1,angle=0]{figures/arrowWhite.eps}}\end{overpic}
	\begin{overpic}[width=\textwidth,percent,angle=270,origin=c]{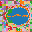}\end{overpic}
\end{subfigure}
\begin{subfigure}[c]{0.116\textwidth}
    \begin{minipage}[t][0.7\baselineskip][t]{\textwidth}
		\centering\footnotesize
		Euclidean
	\end{minipage}
	\begin{overpic}[width=\textwidth,percent,angle=270,origin=c]{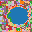}\put(15,30){\includegraphics[scale=1,angle=0]{figures/arrowWhite.eps}}\end{overpic}
	\begin{overpic}[width=\textwidth,percent,angle=270,origin=c]{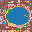}\end{overpic}
\end{subfigure}
\begin{subfigure}[c]{0.116\textwidth}
    \begin{minipage}[t][0.7\baselineskip][t]{\textwidth}
		\centering\footnotesize
		Cosine
	\end{minipage}
	\begin{overpic}[width=\textwidth,percent,angle=270,origin=c]{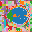}\end{overpic}
	\begin{overpic}[width=\textwidth,percent,angle=270,origin=c]{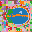}
        \put(65,45){\includegraphics[scale=1,angle=180]{figures/arrowBlack.eps}}
        \put(15,30){\includegraphics[scale=1,angle=0]{figures/arrowWhite.eps}}
    \end{overpic}
\end{subfigure}
\caption{
For two normal and two anomalous images, we visualized the index of the closest component for each patch in terms of component likelihood ($\argmax_k \mathcal{N}(y_n | \mu_k, \Sigma_k)$), Euclidean distance ($\argmin_k ||y_n - \mu_k||_2$), and cosine similarity ($\argmax_k y_n^{\top} \mu_k / (||y_n||_2 \cdot ||\mu_k||_2)$).
To highlight the consistency over multiple samples, we selected samples showing similar anatomical structures for this visualization, with and without lesions. 
Different brain structures of the normal images are seemingly represented by different components: tissues at the centerline by the orange component (white arrow), ventricles by the bright blue component (black arrow).
For the anomalous images, we can observe a similar assignment when using cosine similarity.
However, the component likelihood and the Euclidean distance fail to assign these structures to the correct component in several anomalous cases.
}
\label{fig:meth:component_selection}
\end{figure}

Given the DPMM fitted to the normal distribution of patch embeddings, we can evaluate the alignment of a test sample with the modeled distribution using several metrics.
Specifically, we consider the likelihood of the most probable component as well as the cosine similarity and Euclidean distance to the nearest component center.
As shown in Fig.~\ref{fig:meth:component_selection}, certain components tend to model specific anatomical structures.
Among the three metrics, cosine similarity provides the most consistent assignment of anatomical structures across different samples, particularly for anomalous samples.
Using the full likelihood of the model in Eq.~(\ref{eq:meth:likelihood}) is not an option, as in our experiments the full likelihood is usually dominated by the likelihood of only one component.
Based on this observation, we 1) choose to normalize the embeddings~$y$ for training and testing and 2) use cosine similarity as our anomaly score function.

During fitting the DPMM, some components can vanish as their weight~$\pi_k$ shrinks towards zero.
To suppress all components with vanishing weights, we apply a threshold~$t_{\pi}$ to the component weights first, and the anomaly score function becomes
\begin{equation}
s(y_n) = \max_{k: \pi_k > t_{\pi}} \frac{y_n^{\top} \mu_k}{||y_n||_2 ||\mu_k||_2}.
\end{equation}
Since DINOv2 only creates embeddings of image patches, this yields a coarse, patch-wise anomaly score map.
To obtain pixel-wise anomaly scores, we interpolate the patch-wise anomaly scores to full resolution.

To create an anomaly segmentation mask, we threshold the pixel-level anomaly score map.
To stay in line with the unsupervised setting, we select the threshold such that a predefined false positive rate on the healthy validation set is achieved as in \cite{chen2020unsupervised}.

\section{Experiments and Results}
\subsection{Experimental Setup}
\subsubsection{Model Architecture}
As backbone, we use the small architecture of DINOv2 with the official pretrained weights\footnote{\url{https://github.com/facebookresearch/dinov2} under the Apache 2.0 License.}, leading to a 384-dimensional embedding space.
We truncate the DPMM to 500 components and use diagonal covariance matrices for the Gaussian components.
We set the discount factor in the moving average~$\gamma_t$ to 0.2 to accumulate the statistics over many batches and the component weight threshold~$t_{\pi}$ to \num{1e-6}.
We train the DPMM with a batch size of 12~images, which are resized to a resolution of $448\times448$, resulting in batches of \num{12288}~embedding vectors.
We trained the DPMM for 40~epochs and selected the model achieving the highest log-likelihood on the normal validation samples.

\subsubsection{Datasets}
We conducted our experiments using the anomaly segmentation data\-sets from the BMAD benchmark~\cite{Bao2024BMAD}, which include BraTS2021~\cite{Baid2021BraTS2021Benchmark} (brain MRI), BTCV~\cite{Landman2015BTCV} and LiTs~\cite{Bilic2023LiTs} (liver CT), and RESC~\cite{Hu2019RESC} (retinal OCT).
BraTS2021 comprises \num{8179} normal and \num{3119} anomalous images, the liver CT datasets have \num{2468} normal and \num{733} anomalous images, and RESC contains \num{5408} normal and \num{809} anomalous images.

\subsubsection{Baselines}
We compare our work with multiple unsupervised anomaly detectors, namely DRAEM~\cite{Zavrtanik2021DRAEM}, UTRAD~\cite{Chen2022UTRAD}, RD4AD~\cite{Deng2022RD4AD}, STFPM~\cite{Wang2021STFPM}, PaDiM~\cite{Defard2021PaDiM}, PatchCore~\cite{Roth2022PatchCore}, CFA~\cite{Lee2022CFA}, CFLOW~\cite{Gudovskiy2022CFLOW-AD} and AnomalyDINO~\cite{Damm2024AnomalyDINO}.
For PatchCore, we applied the proposed reduction to 10\% of the memory bank as well as a much stronger reduction to 1024 and to 150 prototypes, which we refer to as few-shot setting.
For AnomalyDINO, we tested the full-shot and one-shot setting, with the one-shot setting corresponding to 1024 prototypes.

\subsubsection{Evaluation Metrics}
We compare our method with state-of-the-art anomaly detectors using the area under the receiver operator characteristic curve~(AUROC) and the area under the precision recall curve (AUPR) at pixel-level.
We also report the per-sample runtime and maximum memory utilization for the anomaly detection stage, measured with a batch size of 1 and one separate thread for data loading using an NVIDIA A6000 GPU with \SI{48}{\giga\byte} of VRAM.
Further, we compare the anomaly segmentation performance with AnomalyDINO using the Dice score at 1\%, 5\%, and 10\% false positive rate on the normal validation images.

To test the statistical significance of our results over PatchCore with 150~prototypes, we run a paired permutation test on the image-wise AUROC and AUPR scores with \num{10000} permutations for the BraTS2021 and the RESC datasets.
Using image-wise scores is necessary as the anomaly score maps for PatchCore and our method are constructed using Euclidean and cosine distance to the nearest prototypes, respectively, and thus cannot be compared directly.

\begin{table}[h!tb]
\centering
\caption{
    AUROC and AUPR scores for pixel-level anomaly detection as well as per-sample runtime (without data loading or metric computation) and memory utilization of the anomaly detection on the test set.
    The methods with the suffix "R50" use WideResNet50~\cite{Zagoruyko2016WideResNet} as the backbone instead of ResNet18~\cite{He2016ResNet}.
    PatchCore-R50 in the standard setting is not available for the BraTS2021 dataset as the coreset subsampling could not fit into the GPU memory.
    We averaged AUROC and AUPR scores for AnomalyDINO in the one-shot setting over 25 runs.
}
\label{tab:res:sota_auc}
\resizebox{\columnwidth}{!}{
\begin{threeparttable}
\begin{tabular}{|L{2.4cm}|cccc|cccc|cccc|}\hline
	                        & \multicolumn{4}{c|}{BraTS2021} & \multicolumn{4}{c|}{BTCV+LiTs} & \multicolumn{4}{c|}{RESC} \\ \hline
	  Method                  & \rotatebox{90}{AUROC [\%]} & \rotatebox{90}{AUPR [\%]} & \rotatebox{90}{Runtime [ms]} & \rotatebox{90}{Memory [GB]} & \rotatebox{90}{AUROC [\%]} & \rotatebox{90}{AUPR [\%]} & \rotatebox{90}{Runtime [ms]} & \rotatebox{90}{Memory [GB]} & \rotatebox{90}{AUROC [\%]} & \rotatebox{90}{AUPR [\%]} & \rotatebox{90}{Runtime [ms]} & \rotatebox{90}{Memory [GB]} \\ \hline
    DRAEM                   & 92.70 & 38.66 & 149 &  0.5 & 93.82 & 16.01 & 125 & 0.5 & 86.78 & 31.86 & 130 &  0.5  \\
    UTRAD                   & 92.81 & 24.63 &  43 &  0.9 & 84.28 &  1.25 &  45 & 0.9 & 55.52 &  6.00 &  42 &  0.9  \\
    RD4AD                   & 97.81 & 52.11 & 139 &  0.4 & 94.35 &  3.48 & 118 & 0.4 & 96.92 & 67.56 & 117 &  0.4  \\
    STFPM                   & 96.64 & 45.45 & 122 &  0.0 & 94.70 &  3.20 & 106 & 0.0 & 93.85 & 45.07 & 108 &  0.0  \\
    STFPM-R50               & 97.32 & 52.87 & 133 &  0.2 & 96.14 &  5.10 & 118 & 0.2 & 94.78 & 52.66 & 111 &  0.2  \\
    PaDiM                   & 85.77 &  8.21 & 145 &  0.2 & 89.77 &  1.66 & 127 & 0.2 & 90.00 & 31.07 & 124 &  0.2  \\
    PaDiM-R50               & 83.36 &  6.89 & 158 &  3.7 & 90.48 &  1.79 & 131 & 3.7 & 92.12 & 35.31 & 129 &  3.7  \\
    PatchCore               & 98.34 & 67.37 & 202 &  3.4 & 96.67 &  5.66 & 132 & 0.7 & 97.08 & 70.47 & 150 &  2.0  \\
    PatchCore-R50           & \multicolumn{4}{c|}{N/A}   & 96.63 &  5.99 & 160 & 1.9 & 97.55 & 72.62 & 218 &  5.0  \\
    CFA                     & 97.27 & 54.15 & 134 &  0.3 & 97.29 &  7.65 & 125 & 0.3 & 91.65 & 26.52 & 116 &  0.3  \\
    CFLOW                   & 95.18 & 32.63 & 255 &  1.0 & 97.13 & 11.63 & 223 & 0.9 & 94.79 & 53.21 & 219 &  0.9  \\
    AnomalyDINO             & 97.71 & 58.69 & 422 & 33.1 & 97.24 & 17.91 & 107 & 6.9 & 94.04 & 50.45 & 984 & 38.1  \\\hline
    PatchCore-R50\tnote{1}  & 95.38 & 55.07 & 149 &  0.2 & 45.70 &  0.32 & 130 & 0.1 & 93.28 & 61.88 & 132 &  0.1  \\
    PatchCore-R50\tnote{2}  & 95.02	& 42.68	& 158 &  0.2 & 16.31 &  0.23 & 139 & 0.1 & 78.70 & 25.98 & 136 &  0.1  \\
    \multicolumn{1}{|r|}{$\blacktriangle$} & 96.57 & 44.74 & && 98.17 & 14.30 & && 92.24 & 42.63 & &\\
    AnomalyDINO\tnote{1*}   & 94.61 & 31.21 &  24 &  0.2 & 94.60 &  5.98 &  26 & 0.2 & 89.52 & 36.98 &  49 &  0.4  \\
    \multicolumn{1}{|r|}{$\blacktriangledown$} & 84.97 & 8.72 & && 89.45 & 2.93 & && 85.10 & 26.51 & &\\
    Ours                    & 96.21\tnote{\textdagger} & 43.43\tnote{\textdagger} &  39 &  1.6 & 95.46 & 9.02 &  35 & 1.5 & 90.20\tnote{\textdagger} & 41.66\tnote{\textdagger} &  58 &  2.0  \\\hline
\end{tabular}
\begin{tablenotes}
    \item [1] few-shot with 1024 prototypes
    \item [2] few-shot with 150 prototypes
    \item [*] max ($\blacktriangle$), average, and min ($\blacktriangledown$) value over all seeds provided
    \item [\textdagger] statistically significant improvement over PatchCore-R50\textsuperscript{2} (permutation test with $p < 1\%$)
\end{tablenotes}
\end{threeparttable}
}
\end{table}

\subsection{Results}
\begin{figure}[tb]
\centering
\includegraphics[width=\textwidth]{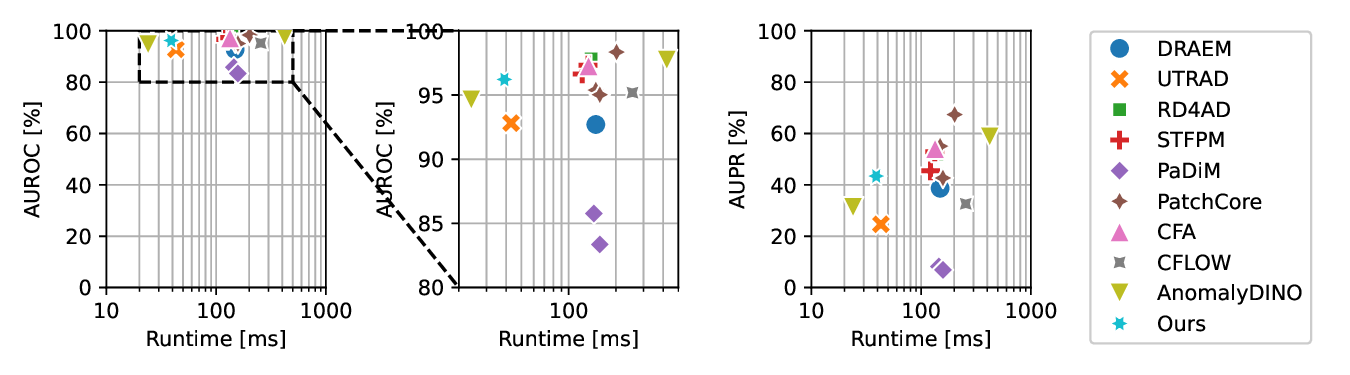}
\caption{
Visualization of the trade-off between runtime and anomaly segmentation performance on the BraTS2021 dataset.
Methods with high anomaly segmentation performance and low runtime are at the top left in each plot.
}
\label{fig:res:runtime_vs_score}
\end{figure}

We provide a quantitative comparison of our method with state-of-the-art approaches in Tab.~\ref{tab:res:sota_auc}.
Over all datasets, only AnomalyDINO in the full-shot setting manages to outperform our approach consistently.
However, AnomalyDINO comes with enormous memory requirements and with long runtimes.
For BraTS and BTCV+LiTs, the AUROC for our approach is only 2.13\% and 1.83\% lower than for the best-performing method, respectively.
With the exception of the one-shot AnomalyDINO on the RESC dataset, our method runs more than 1.8~times faster than all better-performing methods.
This trade-off is also visualized in Fig.~\ref{fig:res:runtime_vs_score}.
Our method is one of the furthest towards low runtimes and high AUROC and AUPR scores, indicating a better trade-off than state-of-the-art methods.

Our approach ends up with roughly 120 to 150~prototypes, while the remaining components vanish.
Comparing with similarly sized coresets for PatchCore or the one-shot setting for AnomalyDINO, our approach outperforms both in almost all metrics, while still having a substantially lower runtime compared to PatchCore.
The implementation of AnomalyDINO is more optimized for efficiency than ours, for instance, they use Faiss~\cite{Johnson2019faiss} for the nearest neighbor search and thus reach faster runtimes. Using a similar optimization, we expect our method to reach even lower runtimes.

We report Dice scores for our approach and both settings of AnomalyDINO in Tab.~\ref{tab:res:sota_dice}.
Not surprisingly, AnomalyDINO in the full-shot setting yields the highest scores for most of the thresholds.
However, our approach outperforms the one-shot AnomalyDINO in all but one case, usually by quite some margin.
Despite using only a few prototypes, our method demonstrates its effectiveness in reducing the gap to methods with many prototypes.

\begin{table}[tb]
\centering
\caption{
    Dice scores for anomaly segmentation for three different thresholds selected to achieve a fixed false positive rate (FPR) on the normal validation set.
    Results for the one-shot setting are averaged over 25~samples.
}
\label{tab:res:sota_dice}
\resizebox{\columnwidth}{!}{
\begin{tabular}{|l|ccc|ccc|ccc|}\hline
 & \multicolumn{3}{c|}{BraTS2021} & \multicolumn{3}{c|}{BTCV+LiTs} & \multicolumn{3}{c|}{RESC} \\ 
\multicolumn{1}{|r|}{FPR}                & 10\%  & 5\%   & 1\%   & 10\% & 5\%   & 1\%   & 10\%  & 5\%   & 1\%   \\ \hline
AnomalyDINO (full-shot) & 29.26 & 38.16 & 54.24 & 7.19 & 10.48 & 18.69 & 39.28 & 45.82 & 52.30 \\
AnomalyDINO (one-shot)  & 31.89 & 37.45 & 32.98 & 6.17 &  8.20 & 10.63 & 36.48 & 40.34 & 33.88 \\
Ours                    & 31.64 & 41.45 & 45.87 & 6.19 &  8.52 & 14.76 & 38.00 & 43.74 & 43.49 \\ \hline
\end{tabular}
}
\end{table}

\begin{table}[tb]
\centering
\caption{
    Ablation results for the anomaly score function and for the normalization of the DINOv2 features.
}
\label{tab:res:ablation}
\resizebox{\columnwidth}{!}{
\begin{tabular}{|cc|cc|cc|cc|}\hline
    \multirow{2}{*}{\makecell{Anomaly Score\\Function}} & \multirow{2}{*}{\makecell{Feature\\Normalization}} & \multicolumn{2}{c|}{BraTS2021} & \multicolumn{2}{c|}{BTCV+LiTs} & \multicolumn{2}{c|}{RESC} \\
	 &  & AUROC & AUPR & AUROC & AUPR & AUROC & AUPR \\ \hline
    Likelihood & False & 92.85 & 23.56 & 94.62 & 5.90 & 85.92 & 29.04 \\
    Euclidean  & False & 93.52 & 29.04 & 93.99 & 7.54 & 86.26 & 35.47 \\
    Cosine     & False & 95.06 & 38.41 & 95.30 & 7.96 & 89.78 & 42.52 \\ \hline
    Likelihood & True  & 95.49 & 33.44 & 95.66 & 9.92 & 89.27 & 33.80 \\
    Euclidean  & True  & 96.17 & 42.68 & 95.44 & 9.22 & 90.15 & 41.00 \\
    Cosine     & True  & 96.21 & 43.43 & 95.46 & 9.02 & 90.20 & 41.66 \\ \hline
\end{tabular}
}
\end{table}
\subsection{Ablation Studies}
We analyze the choice of cosine distance over Euclidean distance and the log-like\-lihood for our anomaly metric as well as the choice of normalizing the DINOv2 features.
The resulting AUROC and AUPR scores are in Tab.~\ref{tab:res:ablation}.
Generally, the results using normalized features are higher than for unnormalized features.
The detection with cosine distance leads to the highest scores for most of the experiments, further supporting the findings of our preliminary analysis in Sec.~\ref{sec:meth:test}.

\section{Conclusion}
In this work, we proposed an unsupervised anomaly detection framework that leverages the expressive power of DINOv2 embeddings and models their distribution using a Dirichlet Process Mixture model~(DPMM), which automatically adjusts the number of components to the data at hand.
Experimental results on the BMAD benchmark~\cite{Bao2024BMAD} demonstrate that our method achieves competitive performance while significantly reducing the computational cost.
Currently, the feature extraction with DINOv2 is the main bottleneck, so future research could explore this setup with distilled models to further improve efficiency.

\subsubsection*{Acknowledgements}
This project was funded by the Swiss National Science Foundation (SNF) (205320\_200877).

\subsubsection*{Disclosure of Interests}
The authors have no competing interests to declare that are relevant to the content of this article.

%
%
%
\bibliographystyle{splncs04}
\bibliography{references.bib}
\end{document}